\documentclass{ecai} 
\usepackage{pifont}
\usepackage{microtype}
\usepackage{inconsolata}
\usepackage{latexsym}
\usepackage{amssymb}
\usepackage{amsmath}

\usepackage{amsthm}
\usepackage{booktabs}
\usepackage{enumitem}
\usepackage{graphicx}
\usepackage{color}
\usepackage{multirow}
\usepackage{algpseudocode}
\usepackage{algorithm} 
\usepackage{natbib}
\usepackage{colortbl}
\usepackage{xcolor}
\usepackage{booktabs}
\usepackage{subcaption}
\newcommand{\BibTeX}{B\kern-.05em{\sc i\kern-.025em b}\kern-.08em\TeX}

\newcommand{\methodname}{\textsc{PoT-PTQ}}
\newcommand{\cmark}{\ding{51}} % Check mark
\newcommand{\xmark}{\ding{55}} % Cross mark
\usepackage{pifont}  % For \cmark and \xmark

\allowdisplaybreaks

\begin{document}

%%%%%%%%%%%%%%%%%%%%%%%%%%%%%%%%%%%%%%%%%%%%%%%%%%%%%%%%%%%%%%%%%%%%%%%%

\begin{frontmatter}

%%% Use this command to specify your submission number.
%%% In doubleblind mode, it will be printed on the first page.

\paperid{8421} 

%%% Use this command to specify the title of your paper.

\title{\includegraphics[height=13pt]{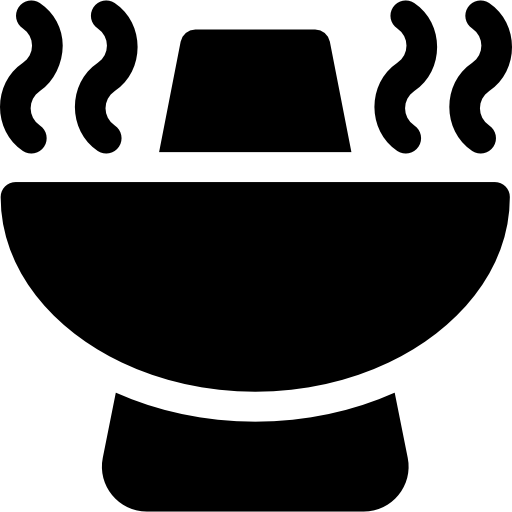}\methodname:\\ A Two-step Power-of-Two Post-training for LLMs}

%%% Use this combinations of commands to specify all authors of your 
%%% paper. Use \fnms{} and \snm{} to indicate everyone's first names 
%%% and surname. This will help the publisher with indexing the 
%%% proceedings. Please use a reasonable approximation in case your 
%%% name does not neatly split into "first names" and "surname".
%%% Specifying your ORCID digital identifier is optional. 
%%% Use the \thanks{} command to indicate one or more corresponding 
%%% authors and their email address(es). If so desired, you can specify
%%% author contributions using the \footnote{} command.

\author[A]{\fnms{Xinyu}~\snm{Wang}\thanks{Corresponding author. Email: \texttt{xinyu.wang5@mail.mcgill.ca}.}}
\author[B]{\fnms{Vahid}~\snm{Partovi Nia}}
\author[C]{\fnms{Peng}~\snm{Lu}}
\author[C,D]{\fnms{Jerry}~\snm{Huang}}
\author[A]{\fnms{Xiao-Wen}~\snm{Chang}}
\author[B]{\fnms{Boxing}~\snm{Chen}}
\author[B]{\fnms{Yufei}~\snm{Cui}\thanks{Corresponding author. Email: \texttt{yufei.cui@huawei.com}.}}

\address[A]{McGill University, Canada}
\address[B]{Huawei Noah's Ark Lab, Canada}
\address[C]{Université de Montréal, Canada}
\address[D]{Mila – Quebec AI Institute, Canada}

%%% Use this environment to include an abstract of your paper.

\begin{abstract}
Large Language Models (LLMs) have demonstrated remarkable performance across various natural language processing (NLP) tasks. However, their deployment is challenging due to the substantial computational resources required. \emph{Power-of-two} (PoT) quantization is a general tool to counteract this difficulty. Albeit previous works on PoT quantization can be efficiently dequantized on CPUs using fixed-point addition, it showed less effectiveness on GPUs. The reason is entanglement of the sign bit and sequential bit manipulations needed for dequantization. 
We propose a novel POT quantization framework for LLM weights that (i) outperforms state-of-the-art accuracy in extremely low-precision number formats, and (ii) enables faster inference through more efficient dequantization. To maintain the accuracy of the quantized model, we introduce a two-step post-training algorithm: (i) initialize the quantization scales with a robust starting point, and (ii) refine these scales using a minimal calibration set.
The performance of our PoT post-training algorithm surpasses the current state-of-the-art in integer quantization, particularly at low precisions such as 2- and 3-bit formats.
Our PoT quantization accelerates the dequantization step required for the floating point inference and leads to $3.67\times$ speed up on a NVIDIA V100, and $1.63\times$ on a NVIDIA RTX 4090, compared to uniform integer dequantization.
\end{abstract}

\end{frontmatter}

\section{Introduction}

Large Language Models (LLMs) have demonstrated remarkable capabilities across a wide range of natural language processing (NLP) tasks, including text generation, summarization, and question answering~\cite{brown2020language,zhang2022opt,team2023gemini,DBLP:journals/corr/abs-2412-15115}. However, deploying them remains challenging due to their memory and computation requirements, posing obstacles when using them in the real world. Quantization has emerged as an effective strategy to reduce these costs by converting full-precision weights into lower-bit representations, e.g. INT8 and INT4, reducing memory usage and accelerating computation. However, aggressive quantization can lead to accuracy degradation, especially in generation tasks, and despite weights being stored in low-bit formats, inference typically involves dequantizing them back to FP16 for general matrix multiplication (GEMM), introducing latency and limiting speed-ups.

Post-training quantization (PTQ)\cite{li2021brecq,lin2024qserve,DBLP:conf/iclr/FrantarAHA23,lin2023awq,shao2024omniquant,llm-int8}, methods offer a practical trade-off: they compress pretrained models without requiring full retraining and are compatible with small calibration sets. While existing PTQ approaches have achieved strong performance in the 3–4 bit regime, maintaining accuracy at greater quantization levels (e.g., 2-bit) remains difficult. Additionally, standard dequantization pipelines often rely on costly mixed-precision operations, further slowing inference. As LLMs scale further, extreme quantization --- such as 2-bit or even binary weights --- becomes increasingly desirable for edge device deployment and low-latency applications. Achieving this level of compression without compromising model quality or inference speed requires new approaches to quantization and hardware efficiency.

Power-of-Two (PoT) quantization offers a promising direction. By constraining weights to signed powers of two, it enables multiplications to be replaced with simple shift-and-add operations, leading to substantial speed gains. Originally explored in computer vision~\cite{yao2022rapq,przewlocka2022power,li2023denseshift}, PoT quantization aligns well with the bell-shaped distributions and exponential tails typically found in trained weight matrices. This structure makes PoT not only hardware-efficient but also statistically well-suited for deep models. However, when applied to LLMs, existing methods fail to retain accuracy due to coarse rounding and lack of effective post-training calibration. Moreover, naive dequantization can be inefficient on modern GPUs due to bit-level dependencies and sign-bit entanglement. Existing attempts to apply power-of-two quantization to LLMs have resulted in significant accuracy degradation, especially at extreme bit-widths (Table~\ref{tab:pot_quantization}).
% \textcolor{red}{The table is missing.}

In this work, we propose a novel post-training quantization framework for LLMs using power-of-two values. Our method achieves both high accuracy in the low-bit regime and fast, hardware-friendly inference.\begin{figure*}[t]
    \centering
    \includegraphics[width=0.34\textwidth,height=40mm]{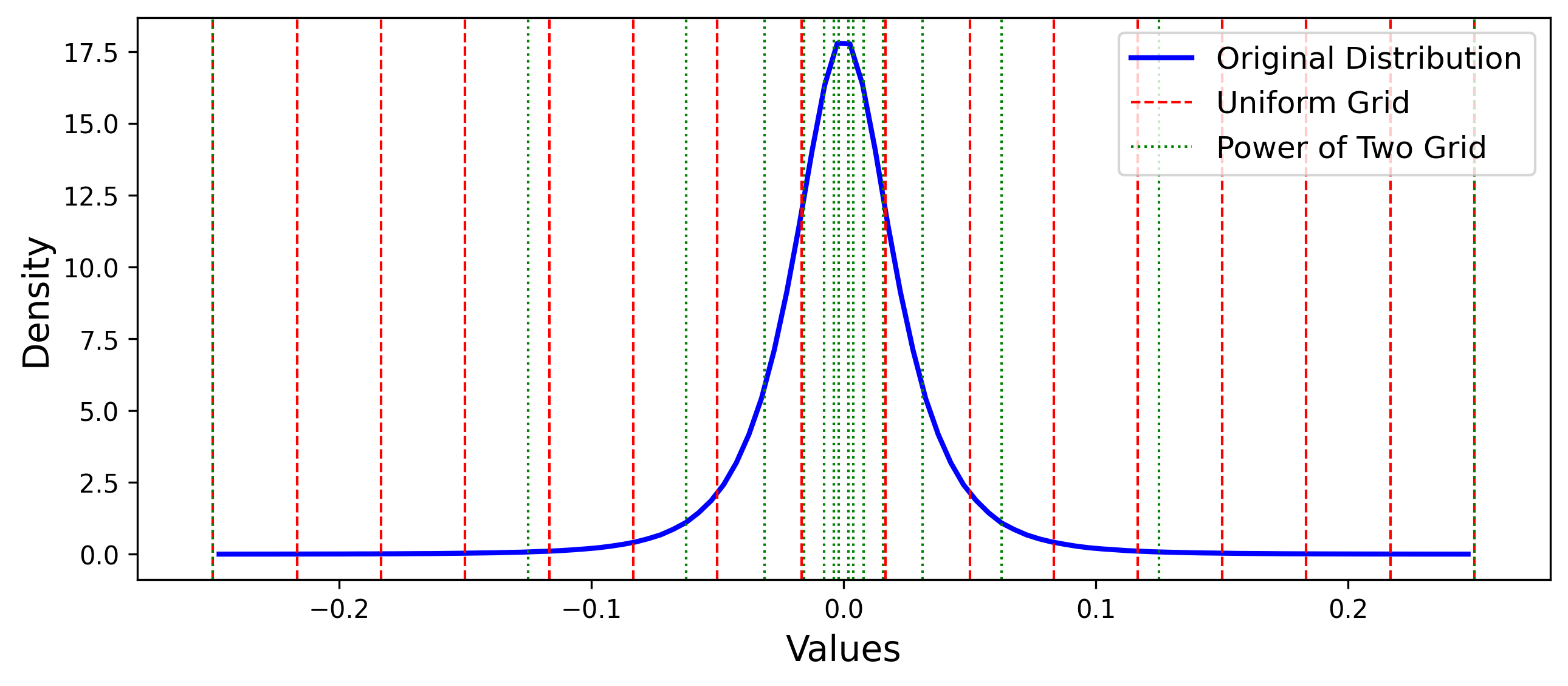}
    \includegraphics[width=0.62\textwidth,height=40mm]{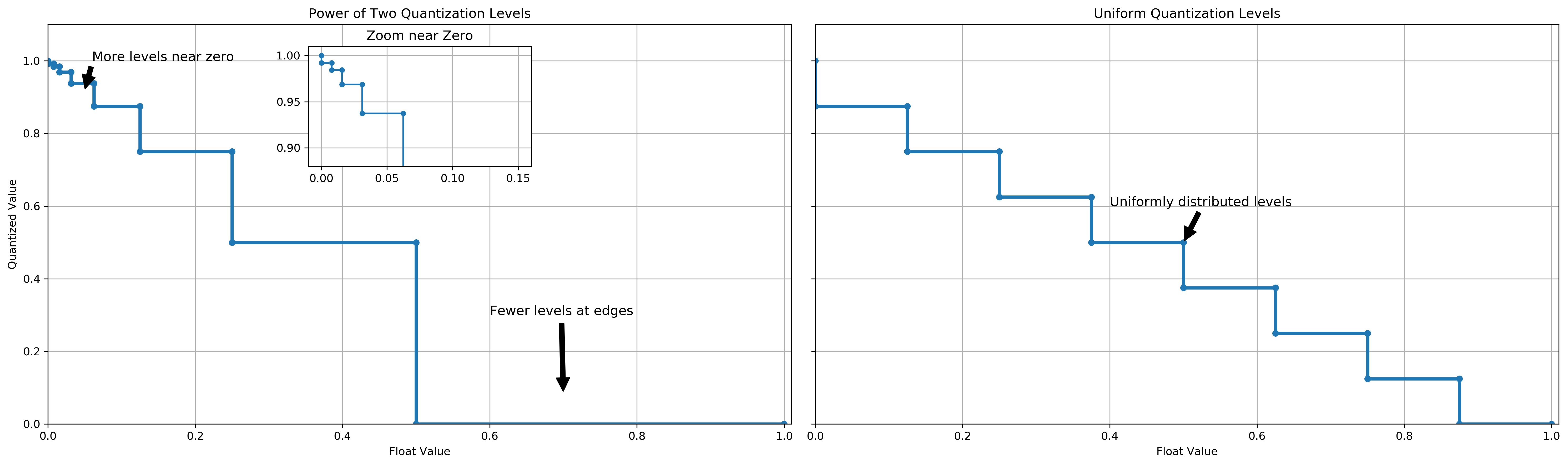}
    \caption{
    Left: Density distribution of the first weight matrix in LLaMA-7B, exemplifying the bell-shaped or exponential decay commonly observed in LLM weight distributions.  
    Middle: Quantization levels for power-of-two (PoT) quantization, showing finer resolution near zero, aligning well with the distribution's high-density region.  
    Right: Quantization levels for uniform quantization, which allocate levels evenly and poorly capture the dense region near zero. \\
    }
    \label{fig:distribution_and_levels}
\end{figure*}
% \textcolor{red}{Do you mean "Our method achieves high accuracy in the low-bit regime while enabling fast, hardware-efficient inference"?}. 
We highlight our main contributions as follows:
\begin{itemize}
    \item We introduce a two-stage post-training algorithm that combines robust scale initialization with lightweight calibration tailored to the PoT structure.
    \item We demonstrate our method consistently outperforms strong PTQ baselines at 2 and 3-bit precision across standard benchmarks.
    \item We develop a GPU-optimized dequantization kernel that leverages bitwise parallelism, resulting in up to $3.67\times$ speed-up on an NVIDIA V100 and $1.63\times$ on a RTX 4090 compared to standard integer dequantization.
\end{itemize}

Our results highlight power-of-two quantization as a scalable and efficient solution for high-performance LLM deployment under stringent hardware constraints.

% \textcolor{red}{Figure 1 is not referred to.}

% \begin{figure}[t]
%     \centering
%     \includegraphics[width=\linewidth]{fig/density_plot_with_grids.png}
%     \caption{Left: Typical LLM weight distribution follows a bell or exponential shape. Right: Comparison of quantization levels in uniform vs. power-of-two quantization. PoT levels align more naturally with the long-tailed distribution, especially in low-bit settings.}
%     \label{fig:distribution_and_levels}
% \end{figure}

\section{Background}

\subsection{Quantization Methods for LLMs}

Recent advances in LLM quantization primarily fall into two categories: \emph{weight-only}, which reduces the bit-width of weights while keeping activations in higher precision, and \emph{weight-activation}, where both components are compressed. Weight-only quantization is particularly attractive for auto-regressive generation as it directly reduces the memory/bandwidth cost during inference while keeping the model forward computation simple and stable.

Several post-training quantization (PTQ) methods have been proposed to enable low-bit inference without retraining. \texttt{LLM.int8}~\cite{dettmers2022gpt3} performs 8-bit quantization with a dedicated outlier channel retained in higher precision. SmoothQuant~\cite{xiao2023smoothquant} applies a per-channel scaling between weights and activations to mitigate outlier effects, and Outlier Suppression~\cite{Wei2023OutlierSA} takes a similar approach by explicitly isolating and handling activation outliers.

Focusing on weight-only PTQ, GPTQ~\cite{DBLP:conf/iclr/FrantarAHA23} proposes a one-shot calibration method based on second-order optimization, achieving strong performance down to 3-bit quantization. AWQ~\cite{lin2023awq} improves robustness via mixed-precision quantization that selectively preserves salient weights in higher precision. OmniQuant~\cite{shao2024omniquant} introduces block-wise learnable clipping bounds to reduce quantization error, while SqueezeLLM~\cite{kim2023squeezellm} explores non-uniform quantization using $k$-means clustering. QuIP~\cite{chee2023quip} refines GPTQ’s optimization and extends support to 2-bit quantization with improved calibration. SpQR~\cite{dettmers2023spqr} and OWQ~\cite{Lee2023OWQOW} further push compression limits by mixing high and low-bit representations based on importance scores.

While these methods enable impressive accuracy at 3-bit and 4-bit regimes, they face challenges at extreme low bit-widths (e.g., 2-bit),
% \textcolor{red}{This implies 1-bit is also one option. Is this the case?}
where discretization artifacts become more prominent. Moreover, many mixed-precision designs introduce dequantization inefficiencies, as multipliers still need to convert integer values to FP16 for matrix multiplication kernels.

In what follows, we turn to power-of-two quantization, which offers a promising direction for reducing both quantization error and dequantization overhead.
\begin{figure*}[t]
    \centering
    \resizebox{0.8\linewidth}{!}{
        \includegraphics[width=\linewidth]{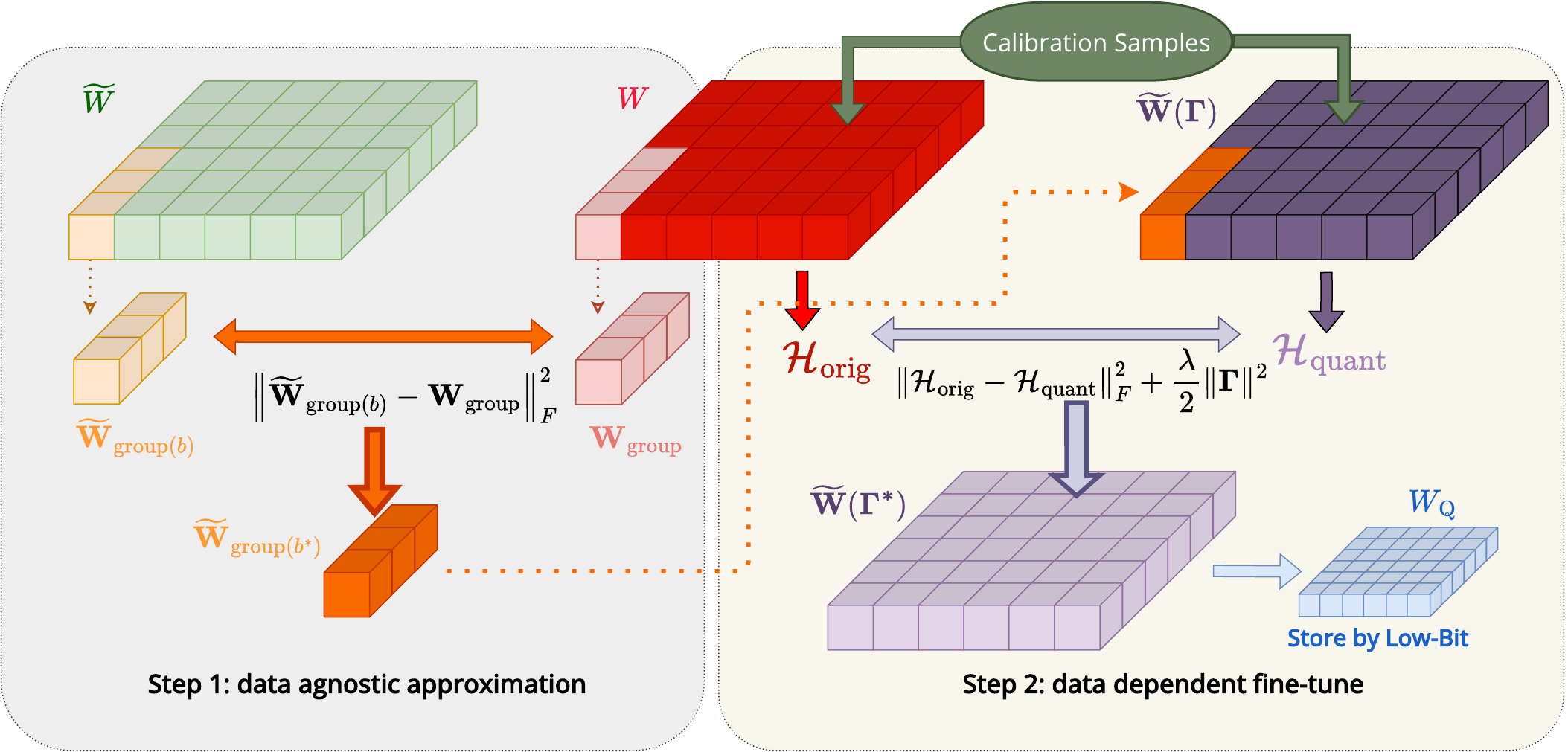}
    }
    \caption{The two step algorithm. \textbf{Step 1}: The scale is adjusted by aligning the weight matrices (left). \textbf{Step 2}: Activations are aligned using minimal calibration data (rights). \\ }
    \label{fig:two_step}
\end{figure*}
% \begin{figure*}[t]
%     \centering
%     \includegraphics[width=\textwidth]{fig/dequant.pdf}
%     \caption{Multiplication between FP16 scale and power-of-two weights for a 3-bit example, implemented through efficient bit manipulation and fixed-point integer addition for accelerated PoT dequantization.}
%     \label{fig:dequant}
% \end{figure*}

\subsection{PoT Quantization}

PoT quantization restricts each weight value to a signed power of two, such as \( \pm 2^{-2}, \pm 2^{-1}, \pm 2^0 \), forming a logarithmically spaced set of quantization levels. Unlike uniform quantization, which allocates evenly spaced intervals, PoT uses a geometric progression that offers both statistical and computational advantages.

\textbf{(1) Statistical alignment with weight distributions.}  
As shown in Figure~\ref{fig:distribution_and_levels} (left), weights in LLMs typically follow bell-shaped or exponentially decaying distributions. The geometric structure of PoT levels aligns more naturally with these distributions—especially in low-magnitude regions—resulting in better approximation under limited bit-widths. In contrast, uniform quantization over-allocates levels to large values and poorly captures the dense region near zero.

\textbf{(2) Hardware-efficient dequantization.}  
PoT quantization enables fast dequantization through shift and add operations. Each quantized weight is reconstructed as \( \pm 2^{\sf E} \cdot S \), where \( S \) is a per-group scaling factor. Since \( 2^{\sf E} \) can be implemented using bit shifts, PoT dequantization is particularly efficient for matrix multiplication on modern GPUs.

Despite these attractive properties, directly applying existing post-training quantization (PTQ) methods \cite{you2020shiftaddnet, DBLP:conf/iclr/LiDW20, elhoushi2021deepshift, li2021s, li2023denseshift} to PoT often results in severe accuracy degradation. Most prior techniques are designed for uniform quantization and do not adapt well to the coarse, non-linear nature of PoT levels. The primary challenge lies in properly determining scaling factors that map weights into the limited exponent range, especially under aggressive bit constraints like 2-bit or 3-bit. Improper scaling leads to large quantization errors that existing activation-based calibration cannot resolve.

These limitations highlight the need for a dedicated PoT quantization framework that explicitly accounts for the structural properties of LLMs and the discrete nature of power-of-two representations. In the following sections, we present such a framework based on principled scale initialization and lightweight calibration.

\section{Two-Step Power-of-Two Post-Training Quantization}
\subsection{Preliminaries}

We focus on weight-only quantization for transformer-based LLMs using Power-of-Two (PoT) representations. Let \( \mathbf{W}^{(l)} \in \mathbb{R}^{d \times d} \) denote the weight matrix at layer \( l \), where each element \( \mathbf{W}_{ij}^{(l)} \) is quantized independently. PoT quantization represents each weight as a signed power-of-two value scaled by a learnable factor:
\begin{equation}
    \widetilde{\mathbf{W}}^{(l)}_{ij} = \mathbf{S}_{ij}^{(l)} \cdot \mathbf{P}_{ij}^{(l)} \cdot 2^{\mathsf{E}_{ij}^{(l)}}, 
    \label{eq:pot_dequant_general}
\end{equation}
where
\begin{itemize}
    \item \( \mathbf{S}_{ij}^{(l)} \in \mathbb{R}_+ \) is the scale shared within a quantization group;
    \item \( \mathbf{P}_{ij}^{(l)} \in \{-1, +1\} \) is the sign bit, defined as \( \mathbf{P}_{ij}^{(l)} = \mathrm{sign}(\mathbf{W}_{ij}^{(l)}) \);
    \item \( \mathsf{E}_{ij}^{(l)} \in \mathbb{N}_0 \) is the quantized exponent, computed as:
    \begin{equation}
        \mathsf{E}_{ij}^{(l)} = \mathrm{clamp}\left( \mathrm{round}\left( \log_2 \left( \left|\mathbf{W}_{ij}^{(l)}\right|/{\mathbf{S}_{ij}^{(l)}} \right) \right), 0, q_{\mathrm{max}} \right),
        \label{eq:pot_exponent}
    \end{equation}
    where \( q_{\mathrm{max}} = 2^{n-1} - 1 \) for \( n \)-bit quantization,
    and the $\mathrm{clamp}$ operation limits a value to a specified interval;
   specifically for real values \( x \), \( a \), and \( b \)
\begin{equation}
    \mathrm{clamp}(x, a, b) = \min(\max(x, a), b).
\end{equation}
\end{itemize}

We can write \eqref{eq:pot_dequant_general} in the matrix form:
%The element-wise dequantized matrix can be expressed compactly as:
\begin{equation}
    \widetilde{\mathbf{W}}^{(l)} = \mathbf{S}^{(l)} \circ \mathbf{P}^{(l)} \circ 2^{\mathsf{E}^{(l)}},
    \label{eq:pot_dequant_matrix}
\end{equation}
where \( \circ \) denotes Hadamard (element-wise) multiplication, and all variables retain their element-wise definitions over the matrix.

To reduce memory and computational costs, we apply \textbf{group-wise quantization}: each column of \( \mathbf{W}^{(l)} \) is divided into fixed-size groups (e.g., 64 or 128 rows), and all weights within a group share the same scale \( \mathbf{S}^{(l)}_{ij} \). This is enforced by:
\begin{equation}
    \mathbf{S}^{(l)}_{ij} = \mathbf{S}^{(l)}_{kj}, \quad 
    \text{if } \left\lfloor i/G \right\rfloor = \left\lfloor k/G \right\rfloor,
    \label{eq:groupwise_scale_constraint}
\end{equation}
where \( G \) is the group size.

This representation forms the foundation of our quantization framework. In the following sections, we describe how to initialize and calibrate the scale factors \( \mathbf{S}^{(l)} \), enabling accurate and efficient PoT quantization for large language models.

\subsection{Step 1: Data-Agnostic Scale Initialization}
\label{subsec:step1}

\begin{figure*}[t]
    \centering
    \includegraphics[width=0.33\textwidth]{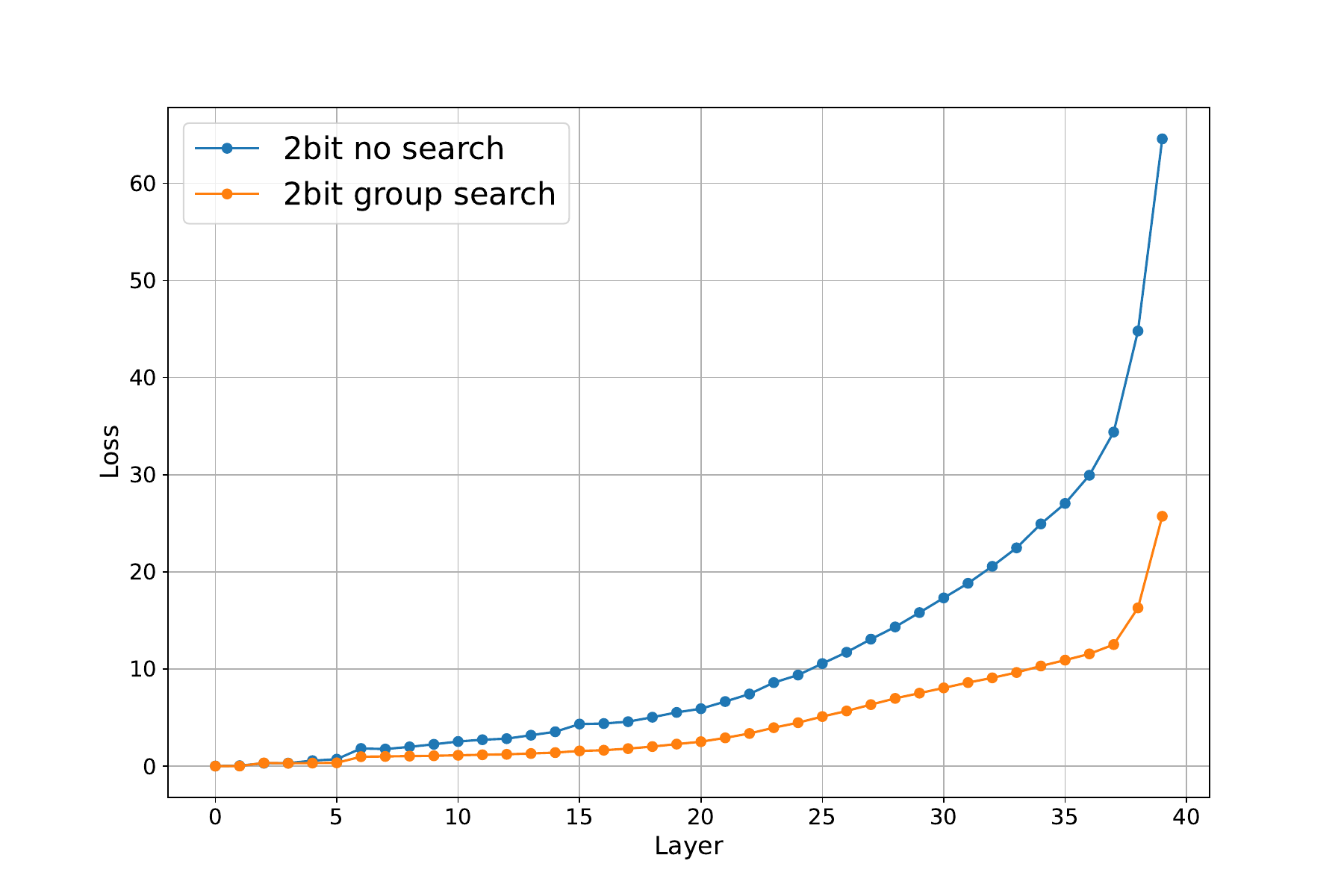}
    \includegraphics[width=0.33\textwidth,height=38mm]{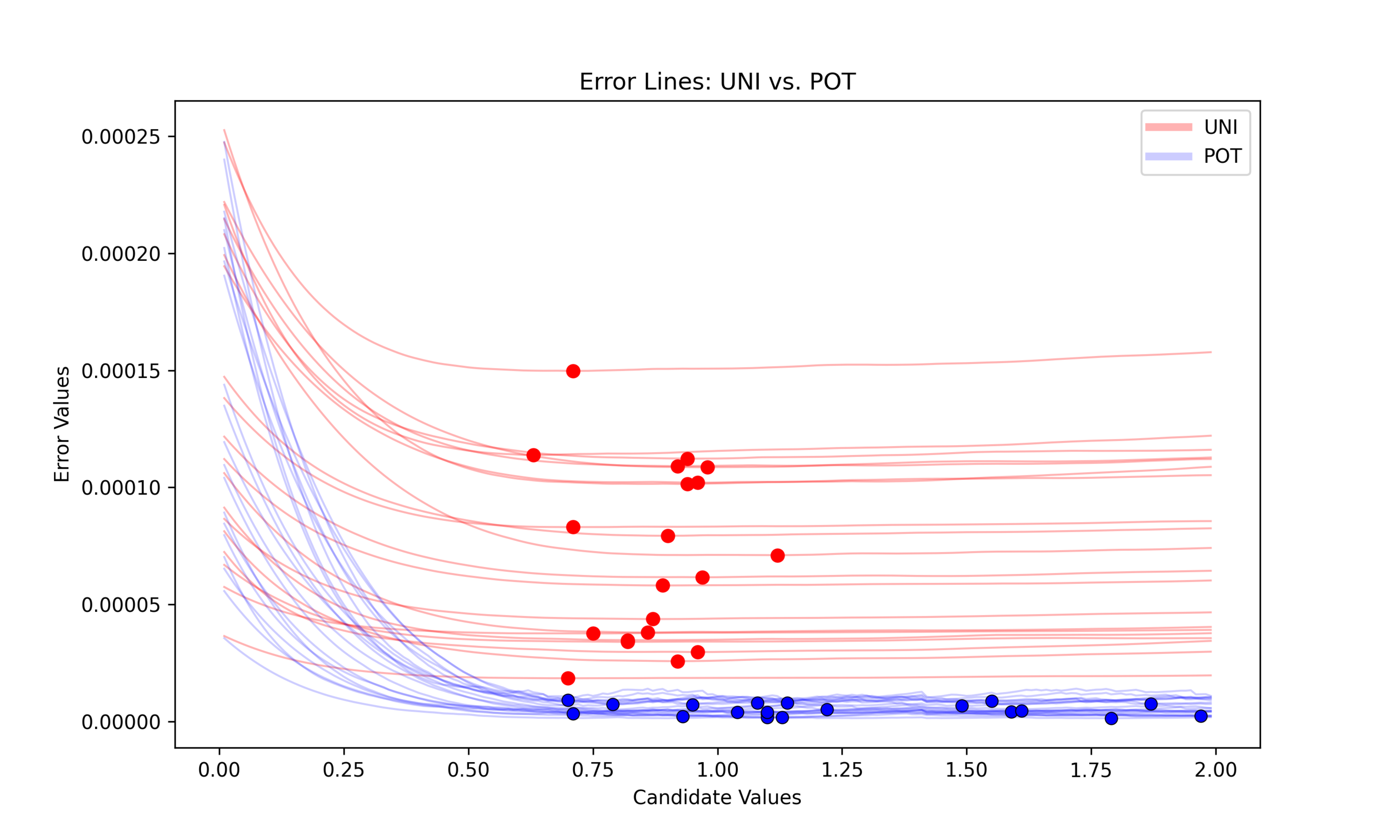}
    \includegraphics[width=0.33\textwidth,height=38mm]{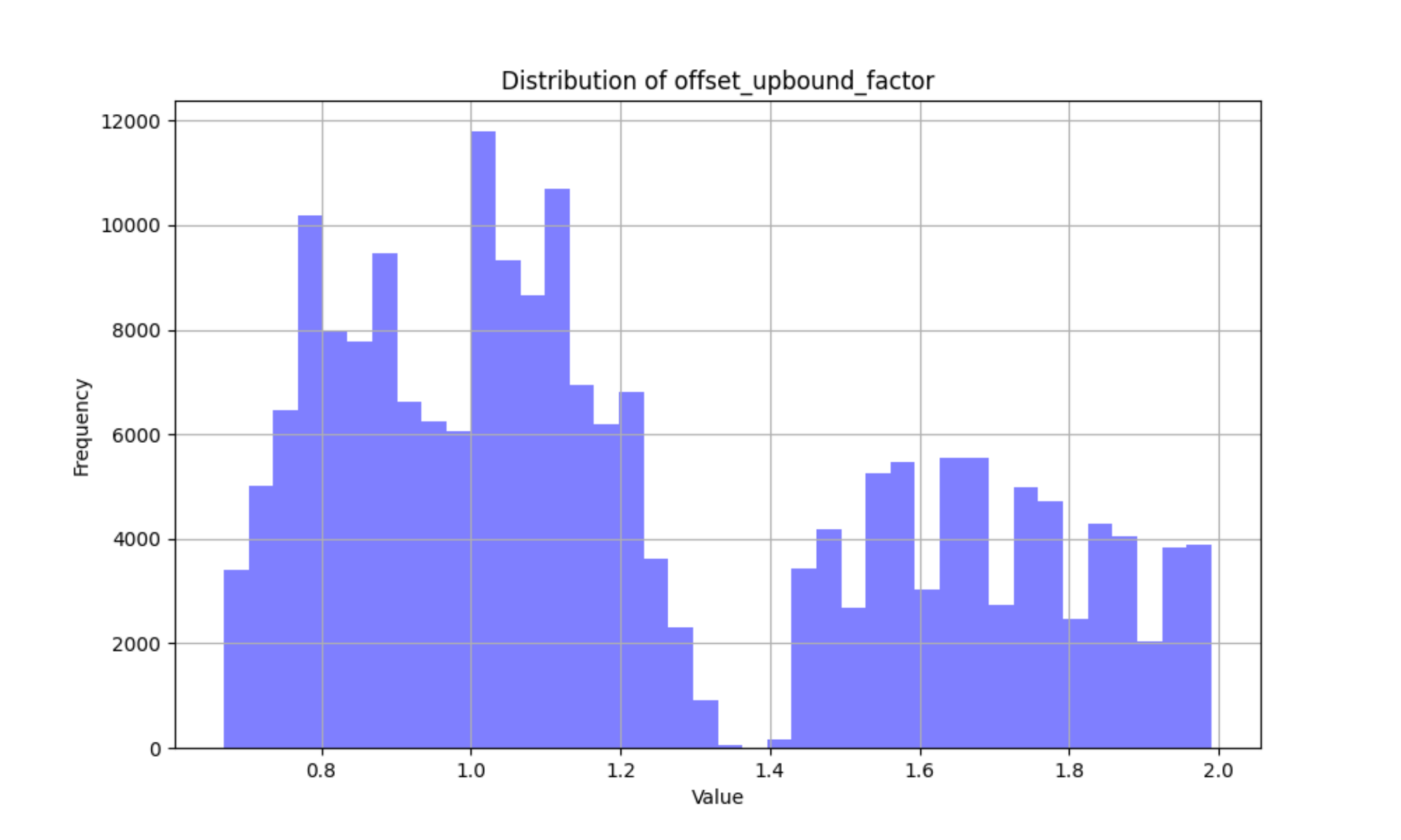}
    \caption{
    Left: Loss in Step1 O/W Step 1   
    Middle: Loss curve of 3-bit PoT quantization shows non-smooth transitions.  
    Right: Histogram of selected scaling multipliers \( b^* \)—many deviate from the naive choice \( b=1 \). \\
    }
    \label{fig:distribution_loss_scale}
\end{figure*}

POT quantization introduces distinct challenges due to its logarithmic and discrete structure. In this stage, we initialize group-wise scaling factors purely from weights, without relying on activation data. This scale initialization is crucial for ensuring quantization robustness under low-bit settings.
\begin{algorithm}[t]
\caption{Parallel Data-Agnostic Scale Initialization (Step 1)}
\label{alg:parallel_initial_scale_search}
\begin{algorithmic}[1]
\For{each weight group \( \mathbf{W}_{\text{group}} \) \textbf{in parallel}}
    \State $s_0 \gets \dfrac{\max |\mathbf{W}_{\text{group}}|}{2^{q_{\text{max}}} - 1}$ \Comment{Initialize base scale}
    \State $\mathcal{B} \gets \{0.01 \cdot i \mid i = 1, \dots, 200\}$ \Comment{Set candidate multipliers}
    \State $Q_1^\text{min} \gets \infty$ \Comment{Initialize minimum error}
    \For{each $b \in \mathcal{B}$}
        \State $s_b \gets s_0 \cdot b$ \Comment{Compute candidate scale}
        \State $\mathsf{E}(b) \gets \mathrm{clamp}\left( \mathrm{round}\left( \log_2\left( \left|\mathbf{W}_{\text{group}}\right|/{s_b} \right) \right), 0, q_{\text{max}} \right)$
        \State $\widetilde{\mathbf{W}}_{\text{group}}(b) \gets s_b \cdot \mathrm{sign}(\mathbf{W}_{\text{group}}) \circ 2^{\mathsf{E}(b)}$
        \State $Q_1(b) \gets \left\| \mathbf{W}_{\text{group}} - \widetilde{\mathbf{W}}_{\text{group}}(b) \right\|_2^2$
        \If{$Q_1(b) < Q_1^\text{min}$}
            \State $Q_1^\text{min} \gets Q_1(b)$
            \State $b^* \gets b$
        \EndIf
    \EndFor
    \State $s^* \gets s_0 \cdot b^*$ \Comment{Optimal scale for this group}
    \State Store $s^*$ into the corresponding position in \( \mathbf{S}^{(l)} \)
\EndFor
\State \textbf{return} \( \mathbf{S}^{(l)} \)
\end{algorithmic}
\end{algorithm}
\paragraph{Motivation: Sensitivity and Non-Smooth Error Surface.}  
PoT quantization results in a highly non-smooth loss landscape, owing to discrete exponent rounding. As illustrated in Figure~\ref{fig:distribution_loss_scale} (middle), small variations in scale lead to abrupt shifts in exponent values, producing sharp error transitions—unlike uniform quantization, which exhibits a smoother loss surface. Furthermore, Figure~\ref{fig:distribution_loss_scale} (right) shows that optimal scales often diverge from the naive initialization (\( b = 1 \)). These properties make gradient-based optimization unreliable and motivate a grid search strategy.

\paragraph{Group Definition.}  
We adopt group-wise quantization, where each column of \( \mathbf{W}^{(l)} \in \mathbb{R}^{\mathbb{R}^{d_{\text{out}} \times d_{\text{in}}}} \) 
% \textcolor{red}{Is the weight matrix at each layer square?}
is partitioned into disjoint groups of size \( G \). Each group contains a contiguous subvector of weights from the same column. Denote such a group as \( \mathbf{W}_{\text{group}} \in \mathbb{R}^{G} \). All elements in a group share the same scaling factor.

\paragraph{Quantization Objective.}  
Given a candidate scale \( s = s_0 \cdot b \), we compute the quantized exponent for each element in the group:
\begin{equation}
    \mathsf{E}_{\text{group}}(b) = \mathrm{clamp}\left( \mathrm{round}\left( \log_2 \left( \frac{|\mathbf{W}_{\text{group}}|}{s_0 \cdot b} \right) \right), 0, q_{\text{max}} \right),
    \label{eq:exponent_group}
\end{equation}
where \( q_{\text{max}} = 2^{n-1} - 1 \) is the maximum exponent level for \( n \)-bit quantization,
and scalars $s_0$ and $b$ are defined later.
We then define the dequantized group reconstruction:
\begin{equation}
    \widetilde{\mathbf{W}}_{\text{group}}(b) = s_0 \cdot b \cdot \mathbf{P}_{\text{group}} \circ 2^{\mathsf{E}_{\text{group}}(b)},
    \label{eq:dequant_group}
\end{equation}
where \( \mathbf{P}_{\text{group}} = \mathrm{sign}(\mathbf{W}_{\text{group}}) \).

The goal is to minimize the mean squared error (MSE) between the original and reconstructed weights:
\begin{equation}
    Q_1(b) = \big\| \mathbf{W}_{\text{group}} - \widetilde{\mathbf{W}}_{\text{group}}(b) \big\|_2^2,
\end{equation}
% \textcolor{red}{The involved quantity is a vector. Why not using the 2-norm instead?}
and select the optimal scale multiplier:
\begin{equation}
    b^* = \arg\min_{b \in \mathcal{B}} Q_1(b), \quad s^* = s_0 \cdot b^*,
    \label{eq:objective_single_group}
\end{equation}
where $\cal B$ is a set of scaling multipliers defined later.

\paragraph{Initial Scale Estimation.}  
We initialize a base scale \( s_0 \) to align the largest magnitude in the group with the top PoT level:
\begin{equation}
    s_0 = \frac{\max |\mathbf{W}_{\text{group}}|}{2^{q_{\text{max}}} - 1}.
\end{equation}

\paragraph{Grid Search Procedure.}  
We evaluate a discrete set of scaling multipliers \( \mathcal{B} = \{ 0.01 \cdot i \mid i = 1, \dots, 200 \} \), and compute \( Q_1(b) \) for each. The best multiplier \( b^* \) and corresponding scale \( s^* \) are retained. This procedure is repeated for every group independently, and the resulting group-wise scales are aggregated into the full matrix \( \mathbf{S}^{(l)} \).

\paragraph{Parallelization.}  
Each weight group is quantized independently, allowing Step 1 to be fully parallelized across GPU threads. This parallelism enables efficient scale initialization even for large models. The procedure is summarized in Algorithm~\ref{alg:parallel_initial_scale_search}, which performs a grid search to minimize reconstruction error per group. A visual overview of Step 1 is shown in \textbf{the left panel of Figure~\ref{fig:two_step}}, where weight matrices are aligned through scale adjustment without any data dependency.

% \textcolor{red}{Confusing. Isn't this algorithm for one group?
% If so, why the for loop in lines 4-18 and why line 19?
% If the algorithm is for all groups, shouldn't lines 1-3 be included in the for loop?
% BTW, is this algorithm referred to in the text?}
\subsection{Step 2: Data-Dependent Fine-Tuning for Layer-Wise Quantization Reconstruction}
\label{subsection:step2}

Following the data-agnostic scale initialization in Step 1 (Section~\ref{subsec:step1}), we introduce a lightweight fine-tuning stage to further refine the group-wise scaling factors \( \mathbf{S}^{(l)} \). This step uses a small calibration set to improve the alignment between the outputs of the quantized and original models. A visual overview of this procedure is shown in the \textbf{right panel of Figure~\ref{fig:two_step}}, where activations are aligned using data-dependent scale adjustments. Importantly, this is achieved by learning a low-dimensional residual parameter \( \boldsymbol{\Gamma} \),
thereby maintaining low training cost and avoiding full model retraining.

\begin{center}
\begin{figure*}[t]
    \includegraphics[width=0.95\textwidth] {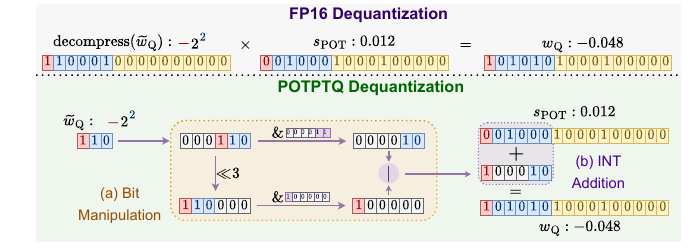}
    \caption{Multiplication between FP16 scale and power-of-two weights for a 3-bit example implemented through efficient bit manipulation and fixed-point integer addition to speed up power of two dequantization. \\ }
    \label{fig:dequant}
\end{figure*}
\end{center}
\vspace{2mm}

\paragraph{Motivation.}
Despite accurate weight reconstruction, the initial scales derived in Step 1 do not directly guarantee output-level consistency due to nonlinear interactions between quantized weights and activation patterns. Fine-tuning with calibration data can mitigate this mismatch. Our method adjusts only the scaling factors via a learnable residual, offering a highly efficient yet effective quantization-aware calibration.

\vspace{2mm}
\paragraph{Optimization Objective.}
Let \( \mathbf{X} \in \mathbb{R}^{B \times T \times d} \) be the input hidden states, where \( B \) is the batch size, \( T \) the sequence length, and \( d \) the hidden dimension. Define the output of the $l$-th transformer block as:
\begin{align}
\mathcal{F}^{(l)}(\mathbf{W}^{(l)}, \mathbf{X}) &= \mathrm{MLP}^{(l)}\left( \mathbf{X} + \mathrm{Attn}^{(l)}(\mathbf{X}) \right), \\
\mathrm{Attn}^{(l)}(\mathbf{X}) &= \mathrm{softmax}\left( \mathbf{Q}^{(l)} \mathbf{K}^{(l)\top}/{\sqrt{d}} \right) \mathbf{V}^{(l)}, \\
\mathrm{MLP}^{(l)}(\mathbf{X}) &= \sigma\left( \mathbf{X} \mathbf{W}_1^{(l)} \right) \mathbf{W}_2^{(l)},
\end{align}
where \( \mathbf{Q}^{(l)} = \mathbf{X} \mathbf{W}_Q^{(l)} \), \( \mathbf{K}^{(l)} = \mathbf{X} \mathbf{W}_K^{(l)} \), and \( \mathbf{V}^{(l)} = \mathbf{X} \mathbf{W}_V^{(l)} \) are the projected queries, keys, and values. \( \sigma(\cdot) \) denotes a nonlinear activation such as GELU.

This formulation also supports grouped-query attention (GQA) \cite{DBLP:conf/emnlp/AinslieLJZLS23}, where queries are projected in more heads than keys and values; our quantization pipeline remains compatible since scale adjustment is performed independently per group.
% \textcolor{red}{Because there is space, Can you write MLP and Attn explicitly?}

The optimization loss is formulated as:
\begin{equation}
    \min_{\boldsymbol{\Gamma}} Q_2(\boldsymbol{\Gamma}) =
    \left\| \mathcal{F}^{(l)}(\mathbf{W}^{(l)}, \mathbf{X}) -
    \mathcal{F}^{(l)}(\widetilde{\mathbf{W}}^{(l)}(\boldsymbol{\Gamma}), \mathbf{X}) \right\|_F^2
    + \frac{\lambda}{2} \| \boldsymbol{\Gamma} \|_F^2,
    \label{eq:optimization_gamma}
\end{equation}
where \( \widetilde{\mathbf{W}}^{(l)}(\boldsymbol{\Gamma}) \) is the reconstructed quantized weight matrix parameterized by the learnable residual \( \boldsymbol{\Gamma} \), and \( \lambda \) controls the strength of the regularization term. The shared scale structure within each quantization group follows Eq.~\eqref{eq:groupwise_scale_constraint}.

\paragraph{Learnable Scale Adjustment.}
We define a refined group-wise scale
\begin{equation}
    \hat{\mathbf{S}}^{(l)}_{ij}(\boldsymbol{\Gamma}) = \mathbf{S}^{(l)}_{ij} \cdot (1 + \boldsymbol{\Gamma}_{ij}),
    \label{eq:adjusted_scale_gamma}
\end{equation}
where \( \boldsymbol{\Gamma}_{ij} \) is initialized to 0 and optimized with gradient-based methods. The dequantization follows:
\begin{align}
\hspace*{-3mm}    
\mathsf{E}_{ij}^{(l)}(\boldsymbol{\Gamma}) &= 
\mathrm{clamp}\left( 
    \mathrm{round}\left( 
        \log_2\left( 
            \frac{|\mathbf{W}^{(l)}_{ij}|}{\hat{\mathbf{S}}^{(l)}_{ij}(\boldsymbol{\Gamma})} 
        \right) 
    \right), 0, q_{\text{max}} 
\right), \label{eq:quantized_exponents_gamma_s} \\
\widetilde{\mathbf{W}}^{(l)}_{ij}(\boldsymbol{\Gamma}) &= 
\hat{\mathbf{S}}^{(l)}_{ij}(\boldsymbol{\Gamma}) \cdot \mathbf{P}_{ij} \cdot 2^{\mathsf{E}_{ij}^{(l)}(\boldsymbol{\Gamma})}.
\label{eq:dequantized_weights_fine_tuned}
\end{align}

% \textcolor{red}{Should $\widetilde{\mathbf{W}}_{\mathrm{Q}}^{(l)}$  
% be replaced by $\mathsf{E}^{(l)}$?}
\vspace{2mm}
\paragraph{Gradient Approximation via STE.}
The rounding operation in Eq.~\eqref{eq:quantized_exponents_gamma_s} is non-differentiable. We apply the \textbf{Straight-Through Estimator (STE)} \cite{DBLP:journals/corr/BengioLC13}:
\begin{equation}
    \frac{\partial \mathsf{E}_{ij}^{(l)}}{\partial \hat{\mathbf{S}}_{ij}^{(l)}}
    \approx
    \frac{\partial}{\partial \hat{\mathbf{S}}_{ij}^{(l)}} 
    \log_2\left( 
        \frac{|\mathbf{W}_{ij}^{(l)}|}{\hat{\mathbf{S}}_{ij}^{(l)}} 
    \right).
\end{equation}
This allows us to backpropagate through the quantization process using the smooth surrogate.

\vspace{2mm}
\paragraph{Efficiency.}
This optimization introduces only one learnable scalar per weight group, and fine-tuning requires only a few epochs on a small calibration set. It is compatible with both SGD and Adam optimizers and scales well to large models without modifying the original model architecture or performing any full-rank retraining.

\vspace{2mm}
\paragraph{Procedure.}
The training procedure is summarized in Algorithm~\ref{alg:optimizing_gamma}.

% \begin{algorithm}[H]
% \caption{Fine-Tuning the Learnable Parameter \( \boldsymbol{\Gamma} \) for Scaling Factors}
% \label{alg:optimizing_gamma}
% \begin{algorithmic}[1]
% \State \textbf{Initialize:} \( \Gamma_{ij} \gets 0 \quad \forall (ij) \)
% \State \textbf{Set Parameters:} Learning rate \( \eta \), weight decay \( \lambda \), epochs \( N \)
% \For{\( \text{epoch} = 1 \) to \( N \)}
%     \For{each calibration batch \( \mathbf{X} \)}
%         \State Compute original output: \( \mathcal{H}_{\text{orig}} \gets \mathcal{F}^{(l)}(\mathbf{W}^{(l)}, \mathbf{X}) \)
%         \State Update scale: \( \hat{\mathbf{S}}_{ij}^{(l)} \gets \mathbf{S}_{ij}^{(l)} \cdot (1 + \Gamma_{ij}) \)
%         \State Quantize exponent: \\ 
%          \qquad \qquad \( \mathsf{E}_{ij}^{(l)} \gets \mathrm{clamp}(\mathrm{round}(\log_2(|\mathbf{W}_{ij}^{(l)}| / \hat{\mathbf{S}}_{ij}^{(l)})), 0, q_{\text{max}}) \)
%         \State Dequantize: \( \widetilde{w}_{ij}^{(l)} \gets \hat{\mathbf{S}}_{ij}^{(l)} \cdot \mathbf{P}_{ij} \cdot 2^{\mathsf{E}_{ij}^{(l)}} \)
%         \State Compute quantized output: \( \mathcal{H}_{\text{quant}} \gets \mathcal{F}^{(l)}(\widetilde{\mathbf{W}}^{(l)}, \mathbf{X}) \)
%         \State Compute loss: \( Q_2 \gets \| \mathcal{H}_{\text{orig}} - \mathcal{H}_{\text{quant}} \|_F^2 + \frac{\lambda}{2} \|\boldsymbol{\Gamma}\|_F^2 \)
%         \State Update \( \Gamma_{ij} \) via gradient descent
%     \EndFor
% \EndFor
% \State \textbf{Return:} Refined scale matrix \( \hat{\mathbf{S}}^{(l)} = \mathbf{S}^{(l)} \circ (1 + \boldsymbol{\Gamma}) \)
% \end{algorithmic}
% \end{algorithm}

\begin{algorithm}[H]
\caption{Fine-Tuning the Learnable Parameter \( \boldsymbol{\Gamma} \) for Scaling Factors}
\label{alg:optimizing_gamma}
\begin{algorithmic}[1]
\State \textbf{Initialize:} \( \boldsymbol{\Gamma}  \gets \mathbf{0}   \)
\State \textbf{Set Parameters:} Learning rate \( \eta \), weight decay \( \lambda \), epochs \( N \)
\For{\( \text{epoch} = 1 \) to \( N \)}
    \For{each calibration batch \( \mathbf{X} \)}
        \State Compute original output: \( \mathcal{H}_{\text{orig}} \gets \mathcal{F}^{(l)}(\mathbf{W}^{(l)}, \mathbf{X}) \)
        \State Update scales: \( \hat{\mathbf{S}}^{(l)} \gets \mathbf{S}^{(l)} 
        \circ (\mathbf{1}\mathbf{1}^\top + \boldsymbol{\Gamma}) \)
        \State Quantize exponent: \\ 
         \qquad \qquad \( \mathsf{E}^{(l)} \gets \mathrm{clamp}(\mathrm{round}(\log_2(|\mathbf{W}^{(l)}| / \hat{\mathbf{S}}^{(l)})), 0, q_{\text{max}}) \)
        \State Dequantize: \( \widetilde{\mathbf{W}}^{(l)} \gets \hat{\mathbf{S}}^{(l)} \circ \mathbf{P} \circ 2^{\mathsf{E}^{(l)}} \)
        \State Compute quantized output: \( \mathcal{H}_{\text{quant}} \gets \mathcal{F}^{(l)}(\widetilde{\mathbf{W}}^{(l)}, \mathbf{X}) \)
        \State Compute loss: \( Q_2 \gets \| \mathcal{H}_{\text{orig}} - \mathcal{H}_{\text{quant}} \|_F^2 + \frac{\lambda}{2} \|\boldsymbol{\Gamma}\|_F^2 \)
        \State Update \( \boldsymbol{\Gamma}  \) via gradient descent
    \EndFor
\EndFor
\State \textbf{Return:} Refined scale matrix \( \hat{\mathbf{S}}^{(l)} = \mathbf{S}^{(l)} \circ (\mathbf{1}\mathbf{1}^\top + \boldsymbol{\Gamma}) \)
\end{algorithmic}
\end{algorithm}
\vspace*{-3mm}
In Line 8, all operations on matrices are carried out elementwise.  
% \textcolor{red}{I have commented out the old version
% and given the new version in the matrix form.
% Is Line 15 needed?}

\section{Efficient Dequantization for PoT Quantized Weights}
\label{sec:efficient_dequantization}

In our proposed quantization framework, we adopt a PoT
dequantization scheme to enable fast and hardware-friendly inference. Unlike traditional uniform quantization that requires floating-point operations, PoT quantization reconstructs weights using bit-shift and integer arithmetic, which are significantly more efficient on modern hardware. This section presents both a conceptual comparison and the detailed implementation of our optimized PoT dequantization method.

\subsection{Power-of-Two vs. Uniform Quantization}

Conventional uniform quantization reconstructs the original weight tensor \( \widetilde{\mathbf{W}}^{(l)} \) as follows:
\begin{equation}
    \widetilde{\mathbf{W}}^{(l)} = (\widetilde{\mathbf{W}}_{\mathrm{Q}}^{(l)} + \mathbf{Z}) \circ \mathbf{S}, \label{eq:dequantized_uniform}
\end{equation}
where \( \widetilde{\mathbf{W}}_{\mathrm{Q}}^{(l)} \) 
are the quantized values,
\( \mathbf{Z} \) is the zero-point offset, and \( S \) is the per-group scaling factor. This approach requires floating-point multiplication and addition during inference, which may incur significant computational latency.

In contrast, PoT quantization reconstructs weights using:
\begin{align}
    \widetilde{\mathbf{W}}^{(l)} &= \mathbf{S}^{(l)} \circ \widetilde{\mathbf{W}}_{\mathrm{Q}}^{(l)}, \\
    \widetilde{\mathbf{W}}_{\mathrm{Q}}^{(l)} &= (-1)^{\sf S} \circ 2^{\sf E}, \label{eq:quantized_weights_pot}
\end{align}
where \( \mathbf{S}^{(l)} \) is a learnable scaling factor and \( \widetilde{\mathbf{W}}_{\mathrm{Q}}^{(l)} \) encodes each quantized weight using a sign bit (\( {\sf S} \)) and an exponent value (\( {\sf E} \)). This encoding allows us to efficiently multiply quantized weights with activations by simply adjusting exponent values, without requiring explicit multiplication.

Each FP16 value \( x \) consists of three parts: the sign bit \( \sf S \), the exponent bits \( \sf E \), and the mantissa bits \( \sf M \):
\[
    x = \overbrace{\sf S}^{\text{Sign}}\ \overbrace{{\sf E}_1, \ldots, {\sf E}_5}^{\text{Exponent}}\ \overbrace{{\sf M}_1, \ldots, {\sf M}_{10}}^{\text{Mantissa}}.
\]
The full FP16 value is computed as:
\begin{equation}
    x = (-1)^{\sf S} \times 2^{{\sf E} - 15} \times 1.{\sf M}.
\end{equation}

Multiplying an FP16 activation \( x \) with a PoT quantized weight \( w \), represented as \( (-1)^{{\sf S'} + {\sf S}} \times 2^{{\sf E} + {\sf E'} - 15} \times 1.{\sf M} \), is computationally efficient due to the exponent additivity property.

\subsection{PoT Dequantization: Bit Manipulation and Integer Addition}
\label{subsec:dequantiztion}
To reconstruct dequantized weights in FP16 format, we design a two-stage dequantization pipeline leveraging:

\begin{enumerate}
    \item \textbf{Bit Manipulation:} Efficiently assembling the signed exponent 283
value from the PoT quantized format. 
    
    \begin{enumerate}
        \item 
     Extract the Exponent Bits:
    Given a 3-bit quantized weight \( w_{\mathrm{Q}} = {\sf S_0}{\sf E_1}{\sf E_2} \), we isolate the exponent portion using an AND mask \texttt{000011}, yielding \( 0000{\sf E}_1{\sf E}_2 \).

    \item Combine Sign and Exponent: 
    We right-shift the quantized bits by three positions and use an AND operation with \texttt{100000} to extract the sign bit. We then use a bitwise OR to combine the sign and exponent into the signed representation \( {\sf S_0}0000{\sf E}_1{\sf E}_2 \).

\end{enumerate}

    \item \textbf{Fixed-Point Integer Addition:}  Efficiently assembling the signed exponent value from the PoT quantized format.
    
    The result is right-shifted by 10 bits to match FP16 exponent alignment and then added to the precomputed FP16 scale \( s \), yielding the final dequantized FP16 weight. This step is carried out in parallel across all quantized values and avoids floating-point operations entirely.
\end{enumerate}
These steps constitute a lightweight, parallelizable pipeline that enables high-throughput inference in GEMM-heavy LLM workloads.

\begin{table}[thbp]
\caption{WikiText-2 perplexity of quantized LLaMA models. Baselines use uniform quant.; PoT uses power-of-two.}
\label{tab:quantization_comparison_wiki}
\centering
\footnotesize
\setlength{\tabcolsep}{3pt}
\begin{tabular}{@{\hskip 1pt}cccccccc@{\hskip 1pt}}
\toprule
\textbf{Bits} & \textbf{Model} & \textbf{Size} & \textbf{RTN} & \textbf{GPTQ} & \textbf{AWQ} & \textbf{OMNI} & \textbf{PoT} \\
\midrule
\multirow{5}{*}{3.25} & \multirow{3}{*}{\texttt{LLaMA1}} & 7B  & 7.01 & 6.55  & 6.46  & 6.16 & \textbf{6.12}  \\
                      &                                  & 13B & 5.88 & 5.62  & 5.51  & 5.46 & \textbf{5.42}  \\
                      &                                  & 30B & 4.87 & 4.80  & 4.63  & 4.58 & \textbf{4.50}  \\
\cmidrule(lr){2-8}
                      & \multirow{2}{*}{\texttt{LLaMA2}} & 7B  & 6.66 & 6.29  & 6.24  & 6.21 & \textbf{6.03} \\
                      &                                  & 13B & 5.51 & 5.42  & 5.32  & 5.28 & \textbf{5.24} \\
\midrule
\multirow{5}{*}{2.25} & \multirow{3}{*}{\texttt{LLaMA1}} & 7B  & $1.90\!\times\!10^3$ & 44.01 & $2.6\!\times\!10^5$ & \textbf{9.77} & 9.79  \\
                      &                                  & 13B & $7.81\!\times\!10^2$ & 15.6  & $2.8\!\times\!10^5$ & \textbf{7.93} & 7.96  \\
                      &                                  & 30B & 68.04 & 10.92 & $2.4\!\times\!10^5$ & 7.13 & \textbf{7.01} \\
\cmidrule(lr){2-8}
                      & \multirow{2}{*}{\texttt{LLaMA2}} & 7B  & $4.20\!\times\!10^3$ & 36.77 & $2.2\!\times\!10^5$ & 11.23 & \textbf{11.03} \\
                      &                                  & 13B & 122.08 & 28.14 & $1.2\!\times\!10^5$ & 8.33 & \textbf{8.29} \\
\bottomrule
\end{tabular}
\end{table}

\section{Experiments and Results}
\label{sec:experiments}
We evaluate \textsc{PoTPTQ}, our proposed two-step \textbf{Power-of-Two (PoT)} quantization framework, for weight-only post-training quantization (PTQ) of large language models. Experiments are conducted on \texttt{\texttt{LLaMA1}}~\cite{touvron2023llama} and \texttt{Llama2}~\cite{touvron2023llama2} models with 7B, 13B, and 30B parameters. The method targets ultra-low precision quantization at 2 and 3-bit levels. All quantization and calibration procedures are performed on a single Tesla V100 GPU (32GB), while kernel benchmarks are additionally run on an RTX 4090 to assess inference efficiency.

\subsection{Experimental Setup}

\textbf{Step 1 (Data-Agnostic Initialization):} A grid search over the interval \([0, 2]\) with step size 0.01 is performed for each quantization group (group size = 128) to identify the optimal initial scale.  
\textbf{Step 2 (Data-Dependent Fine-Tuning):} Using 128 randomly sampled 2048-token sequences from WikiText-2~\cite{merity2016pointer}, we apply light fine-tuning. The learning rate is set to \(1 \times 10^{-3}\), with weight decay \(1 \times 10^{-1}\). Fine-tuning is run for 10 epochs (3-bit) and 40 epochs (2-bit).

\subsection{Perplexity Evaluation}

Table~\ref{tab:quantization_comparison_wiki} presents perplexity results on WikiText-2 across multiple LLaMA model sizes. \textsc{PoTPTQ} consistently achieves lower perplexity than existing PTQ methods, including RTN~\cite{llm-int8}, GPTQ~\cite{DBLP:conf/iclr/FrantarAHA23}, AWQ~\cite{lin2023awq}, and OmniQuant~\cite{shao2024omniquant}, demonstrating its effectiveness for ultra-low-bit quantization.

To evaluate whether existing PTQ methods can directly adopt PoT quantization, we adapt RTN, GPTQ, AWQ, and OmniQuant to operate under power-of-two constraints and report their results in Table~\ref{tab:pot_quantization}. These naïvely adapted methods suffer notable degradation in perplexity, confirming that uniform-to-PoT substitution without proper scale handling leads to suboptimal outcomes. In contrast, \textsc{PoTPTQ} explicitly optimizes scales for PoT representation, yielding significantly better results.

% We further assess generalization on the C4 dataset in Table~\ref{tab:quantization_comparison_C4}. \textsc{PoTPTQ} again outperforms PoT-adapted baselines across all model sizes, confirming the robustness of our method under distribution shift.
\begin{table}[thbp]
\caption{Perplexity of baseline PTQ methods adapted to PoT format versus our PoT method. Naive adaptations degrade performance; our method preserves accuracy.}
\label{tab:pot_quantization}
\centering
\scriptsize
\setlength{\tabcolsep}{3pt} % tighten spacing
\begin{tabular}{@{}c c c 
    >{\raggedright\arraybackslash}p{1.3cm} 
    >{\raggedright\arraybackslash}p{1.3cm} 
    >{\raggedright\arraybackslash}p{1.3cm} 
    >{\raggedright\arraybackslash}p{1.1cm}@{}}
\toprule
\textbf{Avg Bits} & \textbf{Model} & \textbf{Size} & \textbf{AWQ\_POT} & \textbf{GPTQ\_POT} & \textbf{OMNI\_POT} & \textbf{PoT} \\
\midrule
\multirow{5}{*}{3.125} & \multirow{3}{*}{\texttt{\texttt{LLaMA1}}} & 7B  & 6.52 & $8.27\times10^4$ & 6.37 & \textbf{6.25} \\
                       &                                  & 13B & 5.61 & $5.85\times10^4$ & 5.60 & \textbf{5.50} \\
                       &                                  & 30B & 4.72 & $2.61\times10^4$ & 4.75 & \textbf{4.58} \\
\cmidrule(lr){2-7}
                       & \multirow{2}{*}{\texttt{\texttt{Llama2}}} & 7B  & 6.49 & NaN              & 6.46 & \textbf{6.22} \\
                       &                                  & 13B & 5.43 & $6.41\times10^4$ & 5.45 & \textbf{5.34} \\
\midrule
\multirow{5}{*}{2.125} & \multirow{3}{*}{\texttt{\texttt{LLaMA1}}} & 7B  & $2.69\times10^5$ & $2.92\times10^5$ & 888  & \textbf{10.86} \\
                       &                                  & 13B & $2.80\times10^5$ & $1.83\times10^5$ & 487  & \textbf{8.54} \\
                       &                                  & 30B & $2.39\times10^5$ & $1.44\times10^5$ & 297  & \textbf{7.47} \\
\cmidrule(lr){2-7}
                       & \multirow{2}{*}{\texttt{\texttt{Llama2}}} & 7B  & $2.24\times10^5$ & $2.78\times10^5$ & 3730 & \textbf{12.80} \\
                       &                                  & 13B & $1.27\times10^5$ & $1.03\times10^5$ & 812  & \textbf{9.18} \\
\bottomrule
\end{tabular}
\end{table}

\subsection{Harness Evaluation}

Given that \textsc{PoTPTQ} clearly surpasses RTN, GPTQ, and AWQ, and performs competitively with OmniQuant in terms of perplexity, we conduct a downstream evaluation using the Open LLM Leaderboard harness. This benchmark includes a range of QA and reasoning tasks beyond language modeling.

Table~\ref{tab:harness_comparison} compares \textsc{PoTPTQ} and OmniQuant. Our method matches or outperforms OmniQuant on most tasks and achieves higher average performance, indicating that it preserves not only token-level modeling but also broader functional capabilities required in real-world deployments.
\begin{table*}[htbp]
\centering
\caption{Harness evaluation on six tasks comparing PoT and OmniQuant under 3.25-bit and 2.25-bit settings. Bold indicates better performance.}
\label{tab:harness_comparison}
\small
\renewcommand{\arraystretch}{1.1}
\setlength{\tabcolsep}{5.5pt}
\begin{tabular}{llccccccc|ccccccc}
\toprule
\multirow{2}{*}{Model} & \multirow{2}{*}{Method} & \multicolumn{7}{c|}{Avg Bits = 3.25} & \multicolumn{7}{c}{Avg Bits = 2.25} \\
\cmidrule(lr){3-9} \cmidrule(lr){10-16}
& & arc-c & arc-e & boolq & hs & piqa & wg & Avg & arc-c & arc-e & boolq & hs & piqa & wg & Avg \\
\midrule
\multirow{2}{*}{\texttt{LLaMA1} 7B} 
& Omni & 35.6 & \textbf{64.8} & \textbf{71.1} & 53.9 & 77.2 & 64.5 & 61.2  & 26.7 & \textbf{52.1} & 62.2 & \textbf{40.7} & 67.2 & 55.5 & 50.7 \\
& PoT  & \textbf{35.8} & 64.1 & 70.9 & \textbf{54.3} & \textbf{77.5} & \textbf{65.2} & \textbf{61.3} & \textbf{28.1} & 50.1 & \textbf{64.4} & 40.1 & \textbf{67.9} & \textbf{57.3} & \textbf{51.3}\\
\midrule
\multirow{2}{*}{\texttt{LLaMA1} 13B}
& Omni & 39.8 & \textbf{72.7} & \textbf{67.0} & 56.8 & 77.2 & 68.7 & 63.7 & \textbf{31.3} & \textbf{60.1} & 63.1 & \textbf{46.1} & \textbf{72.0} & 61.8 & 55.7 \\
& PoT  & \textbf{40.7} & 71.8 & 65.6 & \textbf{57.0} & \textbf{78.8} & \textbf{70.3} & \textbf{64.0} & 30.1 & 59.5 & \textbf{66.1} & 45.6 & 70.3 & \textbf{62.7} & \textbf{55.7} \\
\midrule
\multirow{2}{*}{\texttt{LLaMA1} 30B}
& Omni & 46.0 & 74.1 & 71.2 & \textbf{61.3} & 79.6 & 74.1 & 67.7  & 32.4 & \textbf{65.6} & 66.1 & 49.9 & 72.3 & \textbf{62.9} & 58.2 \\
& PoT  & \textbf{47.2} & \textbf{73.7} & \textbf{71.6} & 60.8 & \textbf{80.1} & \textbf{75.0} & \textbf{68.1} & \textbf{33.8} & 64.9 & \textbf{66.8} & \textbf{50.2} & \textbf{73.1} & 62.5 & \textbf{58.5} \\
\midrule
\multirow{2}{*}{\texttt{Llama2} 7B}
& Omni & 37.3 & \textbf{67.6} & 71.2 & 54.5 & \textbf{76.5} & 65.7 & 62.1 & \textbf{26.0} & 45.0 & 61.2 & 39.4 & 64.5 & \textbf{54.4} & 48.4 \\
& PoT  & \textbf{38.1} & 66.3 & \textbf{72.0} & \textbf{55.2} & 76.3 & \textbf{66.6} & \textbf{62.4} & 25.8 & \textbf{52.1} & \textbf{63.8} & \textbf{40.3} & \textbf{65.2} & 54.2 & \textbf{50.2} \\
\midrule
\multirow{2}{*}{\texttt{Llama2} 13B}
& Omni & 41.9 & \textbf{72.3} & 69.9 & 57.8 & 78.0 & 67.7 & 64.6 & \textbf{30.0} & \textbf{57.0} & 63.7 & \textbf{44.5} & 68.0 & 53.1 & 52.7 \\
& PoT  & \textbf{42.5} & 70.7 & \textbf{70.1} & \textbf{58.0} & \textbf{79.2} & \textbf{68.3} & \textbf{64.8} & 29.4 & 56.9 & \textbf{68.2} & 43.4 & 68.8 & \textbf{56.9} & \textbf{53.9} \\
\bottomrule
\end{tabular}
\end{table*}

\subsection{Ablation Study}
\label{sec:ablation}

To isolate the contributions of each step in our pipeline, we conduct an ablation study using 2-bit quantization on \texttt{LLaMA1}-7B and \texttt{LLaMA1}-13B. As shown in Table~\ref{tab:model_performance}, Step 1 alone (data-agnostic initialization) achieves reasonable performance, indicating the efficacy of our scale grid search. Applying Step 2 (fine-tuning) without Step 1, however, results in inferior performance, likely due to poor initial scale estimates. The best performance is consistently achieved by combining both steps, confirming their complementary effects: Step 1 provides robust initialization, and Step 2 refines the solution with minimal calibration data.

\subsection{Fine-Tuning Efficiency and Quantization Time}

To evaluate the effectiveness of Step~2, we track the epoch-wise perplexity and loss on WikiText-2 using the LLaMA-13B model. As shown in Table~\ref{tab:llama13b_performance}, both perplexity and loss decrease consistently over 10 epochs with only 128 calibration sequences. This highlights the efficiency of our fine-tuning process, which requires minimal data while achieving steady improvement. By the 10th epoch, the 2-bit model achieves a perplexity of 12.90 and loss of 7.89, demonstrating that our output-level alignment objective (Eq.~\ref{eq:optimization_gamma}) leads to significantly improved model fidelity.

In terms of wall-clock performance, we evaluate the total quantization time required to complete both stages of our framework: Step~1 (data-agnostic grid search for scale initialization) and Step~2 (data-dependent scale refinement via lightweight fine-tuning). All experiments are conducted on a single NVIDIA Tesla V100 GPU with 32GB of memory.

For LLaMA-7B, the complete quantization pipeline takes approximately 0.71 hours ($\sim$43~minutes). This includes exhaustive scale grid search across weight groups and 10 epochs of fine-tuning using a small calibration set.

Notably, the entire quantization pipeline—spanning initialization, calibration, and weight reconstruction—can be executed within a few hours on commodity hardware, without any need for multi-GPU parallelism or model retraining. This makes our framework highly suitable for real-world deployments where turnaround time and hardware constraints are critical considerations.

\subsection{Dequantization Speed}
\label{subsec:dequant_speed_eval}

To measure inference-time efficiency, we benchmark our custom PoT dequantization kernel on both Tesla V100 and RTX 4090 GPUs. As detailed in Section~\ref{subsec:dequantiztion}, the kernel uses integer arithmetic to reconstruct weights efficiently from PoT encodings. Table~\ref{tab:dequant_speed} shows that our kernel achieves a \(3.66\times\) speedup on V100 and a \(1.48\times\) speedup on 4090, compared to standard FP16 dequantization. These results demonstrate that PoT quantization is not only accurate, but also highly efficient for hardware-accelerated inference.

\subsection{Summary}

\textsc{PoTPTQ} delivers state-of-the-art performance for ultra-low-bit weight-only quantization in LLMs. It outperforms existing PTQ baselines under both uniform and PoT formats, generalizes well across datasets, and preserves downstream task capabilities. Furthermore, the approach is scalable, calibration-efficient, and inference-friendly—making it a strong candidate for deployment in real-world, resource-constrained environments.

% \begin{table}[htbp]
% \caption{Perplexity comparison on the C4 dataset under 2.25- and 3.25-bit quantization across LLaMA models. Baselines use uniform quantization; PoT denotes our method.}
% \label{tab:quantization_comparison_C4}
% \centering
% \scriptsize
% \begin{tabular}{@{}cccccccc@{}}
% \toprule
% \textbf{Avg Bits} & \textbf{Model} & \textbf{Size} & \textbf{GPTQ} & \textbf{AWQ} & \textbf{OMNI} & \textbf{RTN} & \textbf{PoT} \\
% \midrule
% \multirow{5}{*}{3.25} 
% & \multirow{3}{*}{\texttt{\texttt{LLaMA1}}} 
% & 7B  & 7.85 & 7.92 & 7.76 & 8.62 & \textbf{7.76} \\
% & & 13B & 7.10 & 7.07 & \textbf{7.06} & 7.49 & 7.05 \\
% & & 30B & 6.47 & 6.37 & 6.38 & 6.58 & \textbf{6.35} \\
% \cmidrule(lr){2-8}
% & \multirow{2}{*}{\texttt{\texttt{Llama2}}} 
% & 7B  & 7.89 & \textbf{7.84} & 7.87 & 8.40 & 7.87 \\
% & & 13B & 7.00 & \textbf{6.94} & 6.99 & 7.18 & 6.98 \\
% \midrule
% \multirow{5}{*}{2.25} 
% & \multirow{3}{*}{\texttt{\texttt{LLaMA1}}} 
% & 7B  & 27.71 & $1.90\times10^5$ & \textbf{13.04} & $1.00\times10^3$ & 14.82 \\
% & & 13B & 15.29 & $2.30\times10^5$ & \textbf{10.40} & 447.64 & 11.62 \\
% & & 30B & 11.93 & $2.40\times10^5$ & \textbf{9.41}  & 99.45  & 11.56 \\
% \cmidrule(lr){2-8}
% & \multirow{2}{*}{\texttt{\texttt{Llama2}}} 
% & 7B  & 33.70 & $1.70\times10^5$ & \textbf{15.45} & $4.90\times10^3$ & 18.87 \\
% & & 13B & 20.97 & $9.40\times10^4$ & \textbf{11.15} & 139.65 & 12.84 \\
% \bottomrule
% \end{tabular}
% \end{table}

\begin{table}[H]
\centering
\caption{Ablation study on the effect of each quantization step for LLaMA models. Step 1 denotes scale initialization; Step 2 denotes calibration-based fine-tuning.}
\label{tab:model_performance}
\small
\renewcommand{\arraystretch}{1.2}
\resizebox{\columnwidth}{!}{  % Auto-fit to page width
\begin{tabular}{c c c r}
\toprule
\textbf{Model} & \textbf{Step 1 (Init)} & \textbf{Step 2 (Tuning)} & \textbf{Perplexity} \\
\midrule
\multirow{4}{*}{\texttt{LLaMA1} 7B}  
& \xmark & \xmark & 408{,}838.25 \\
& \cmark & \xmark & 20{,}135.70 \\
& \xmark & \cmark & 51.87 \\
& \cmark & \cmark & \textbf{9.79} \\
\midrule
\multirow{4}{*}{\texttt{LLaMA1} 13B} 
& \xmark & \xmark & 40{,}328.34 \\
& \cmark & \xmark & 8{,}267.57 \\
& \xmark & \cmark & 103.45 \\
& \cmark & \cmark & \textbf{7.96} \\
\bottomrule
\end{tabular}
}
\end{table}

\begin{table}[htbp]
\centering
\caption{Epoch-wise perplexity and loss on WikiText-2 with \texttt{LLaMA}-13B using PoT quantization. Our output-aligned fine-tuning objective consistently reduces perplexity with only 128 calibration samples.}
\label{tab:llama13b_performance}
\renewcommand{\arraystretch}{1.2}
\small
\setlength{\tabcolsep}{4.5pt} % tighter column padding
\begin{tabular}{c@{\hskip 6pt}c@{\hskip 6pt}c@{\hskip 6pt}c@{\hskip 6pt}c}
\toprule
\textbf{Epoch} & \textbf{3-bit PPL} & \textbf{3-bit Loss} & \textbf{2-bit PPL} & \textbf{2-bit Loss} \\
\midrule
1  & 6.85 & 6.21 & $1.01\!\times\!10^6$ & 25.75 \\
2  & 5.93 & 3.86 & 3209.49              & 20.45 \\
3  & 5.70 & 3.02 & 239.27               & 17.13 \\
4  & 5.59 & 2.54 & 94.25                & 15.39 \\
5  & 5.54 & 2.27 & 49.95                & 13.50 \\
6  & 5.51 & 2.13 & 31.46                & 11.99 \\
7  & 5.49 & 2.04 & 23.30                & 10.55 \\
8  & 5.48 & 1.97 & 18.27                & 9.60  \\
9  & 5.48 & 1.93 & 15.31                & 8.65  \\
10 & \textbf{5.48} & \textbf{1.90} & \textbf{12.90} & \textbf{7.89} \\
\bottomrule
\end{tabular}
\end{table}

% \begin{table}[htbp]
% \centering
% \caption{\texttt{LLaMA} Quantization Time}
% \begin{tabular}{lccc}
% \toprule
% \textbf{Model} & \textbf{7B} & \textbf{13B} & \textbf{30B} \\
% \midrule
% Quantization Time (hours) & 0.71 & 1.26 & 3.07 \\
% \bottomrule
% \end{tabular}

% \label{tab:llama4bit_quantization}
% \end{table}

\begin{table}[H]
\centering
\caption{Dequantization Efficiency: GPU Warp Cycles. Our PoT kernel achieves significantly lower warp latency across architectures.}
\label{tab:dequant_speed}
\begin{tabular}{lccc}
\toprule
\textbf{GPU} & \textbf{Architecture} & \textbf{Uniform (FP16)} & \textbf{PoT (Ours)} \\
\midrule
Tesla V100 & Volta       & 110 & 30 {\footnotesize\color{green!60!black}($3.67\times$ faster)} \\
RTX 4090   & Ada Lovelace & 98  & 60 {\footnotesize\color{green!60!black}($1.63\times$ faster)} \\
\bottomrule
\end{tabular}
\end{table}

\section{Conclusion}

We presented a novel post-training quantization framework that leverages power-of-two (PoT) representations to enable efficient and hardware-friendly inference. Our method introduces a two-stage algorithm consisting of data-agnostic scale initialization and data-driven fine-tuning, effectively addressing the accuracy limitations commonly observed in traditional PoT quantization schemes.

Through comprehensive experiments on LLaMA models at 2-bit and 3-bit precision, we demonstrate that our approach consistently outperforms existing PTQ methods in both perplexity and real-world deployment scenarios. In addition to its strong accuracy, our framework supports parallelizable grid search and lightweight calibration, making it practical for deployment on standard hardware without requiring retraining or large calibration datasets.

Furthermore, our optimized integer-only dequantization kernel significantly accelerates inference, achieving up to $3.67\times$ speedup on NVIDIA V100 and $1.63\times$ on RTX 4090 compared to traditional FP16-based methods. These results highlight the potential of PoT quantization as a scalable and effective solution for low-latency LLM deployment.

\begin{ack}
We gratefully acknowledge Xinlin Li (Microsoft), Alireza Ghaffari (Cerebras), Ali Edalati (Cohere), and Adrian Zhao (University of Toronto) for their helpful discussions. We especially thank Adrian Zhao and Alireza Ghaffari for their assistance with CUDA kernel design and benchmarking, and Xinlin Li for originally proposing the idea of power-of-two quantization.
\end{ack}

%%%%%%%%%%%%%%%%%%%%%%%%%%%%%%%%%%%%%%%%%%%%%%%%%%%%%%%%%%%%%%%%%%%%%%%%
% \bibliographystyle{plain}
%%% Use this command to include your bibliography file.
\bibliography{potpot.bib}

\end{document}